\documentclass[11pt]{article}

\usepackage[preprint]{acl}

\usepackage{times}
\usepackage{latexsym}
\usepackage{amsmath,amssymb}

\usepackage[T1]{fontenc}
\usepackage[utf8]{inputenc}

\usepackage{microtype}

\usepackage{inconsolata}

\usepackage{graphicx}
\usepackage{booktabs}
\usepackage{colortbl}
\usepackage{multirow}
\usepackage{xspace}
\newcommand{\ours}{AgenticRoboRL\xspace}
\definecolor{refcolor}{HTML}{E11D74}
\AtBeginDocument{\hypersetup{linkcolor=refcolor}}
\definecolor{logoorange}{HTML}{F76707}
\definecolor{logonavy}{HTML}{1C2A45}
\definecolor{logoblue}{HTML}{1C7ED6}
\definecolor{logogreen}{HTML}{37B24D}
\usepackage{fontawesome5}
\newcommand{\codeurl}{https://github.com/L-CodingSpace/AgenticRoboRL}
\newcommand{\pageurl}{https://l-codingspace.github.io/AgenticRoboRL/}

\title{\makebox[0pt][r]{\raisebox{-0.15\height}{\includegraphics[height=1.2cm]{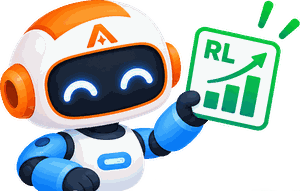}}\hspace{0.3em}}{\huge\texttt{\textcolor{logoorange}{A}\textcolor{logonavy}{gentic}\textcolor{logoblue}{Robo}\textcolor{logogreen}{RL}}:} \\ Bringing Coding-Agent Post-Training to Agentic Robotics}

\author{\textbf{Xinxin Liu}\textsuperscript{1} \quad \textbf{Bingxin Xu}\textsuperscript{2} \quad \textbf{Wenrui Bao}\textsuperscript{1} \quad \textbf{Ming Li}\textsuperscript{1} \\ \textbf{Wentao Bao}\textsuperscript{3} \quad \textbf{Chen Chen}\textsuperscript{1} \quad \textbf{Yuzhang Shang}\textsuperscript{1} \\ \textsuperscript{1}University of Central Florida \quad \textsuperscript{2}University of Southern California \quad \textsuperscript{3}Michigan State University \\ \raisebox{-0.1em}{\faGithub}\ \href{\codeurl}{\textbf{Code}} \quad \raisebox{-0.1em}{\faGlobe}\ \href{\pageurl}{\textbf{Project page}}}

\usepackage{mathtools}
\makeatletter
\g@addto@macro\normalsize{%
  \setlength\abovedisplayskip{5pt plus 2pt minus 3pt}%
  \setlength\belowdisplayskip{5pt plus 2pt minus 3pt}%
  \setlength\abovedisplayshortskip{2pt plus 1pt minus 1pt}%
  \setlength\belowdisplayshortskip{4pt plus 2pt minus 2pt}}
\makeatother

\AtBeginDocument{\setlength{\parfillskip}{0pt plus 150pt}\setlength{\emergencystretch}{1em}}

\begin{document}
\maketitle

\begin{abstract}
Agentic robotics systems combine planners with robot harnesses to perform complex manipulation tasks. Recent work improves these systems through better harnesses and external memory, but relies on powerful closed-source models as planners and does not improve the planners' own capabilities. In contrast, coding agents have improved substantially by post-training the model itself. We therefore investigate whether post-training can also improve robot planners. However, applying it across robot harnesses requires accommodating different execution mechanisms, including VLA skill invocation and control-code generation. To this end, we introduce \texttt{\textbf{AgenticRoboRL}}, a framework for post-training robot planners across different harnesses. The training recipe combines supervised fine-tuning with multi-turn reinforcement learning. To accommodate different execution mechanisms, harness-specific adapters connect recorded planner interactions with task outcomes, enabling a common RL training pipeline while keeping execution components fixed. On LIBERO-PRO, \ours raises the success rate of a 9B-parameter planner from \textbf{18.1\% to 71.0\%} without external memory, approaching the 72.1\% achieved by Codex (GPT-5.5) with external memory. These results demonstrate post-training as an effective approach for improving robot planners. We will release the framework and trained planner checkpoints.
\end{abstract}

\section{Introduction}
\label{sec:intro}

\begin{figure*}[t]
\centering
\includegraphics[width=\textwidth]{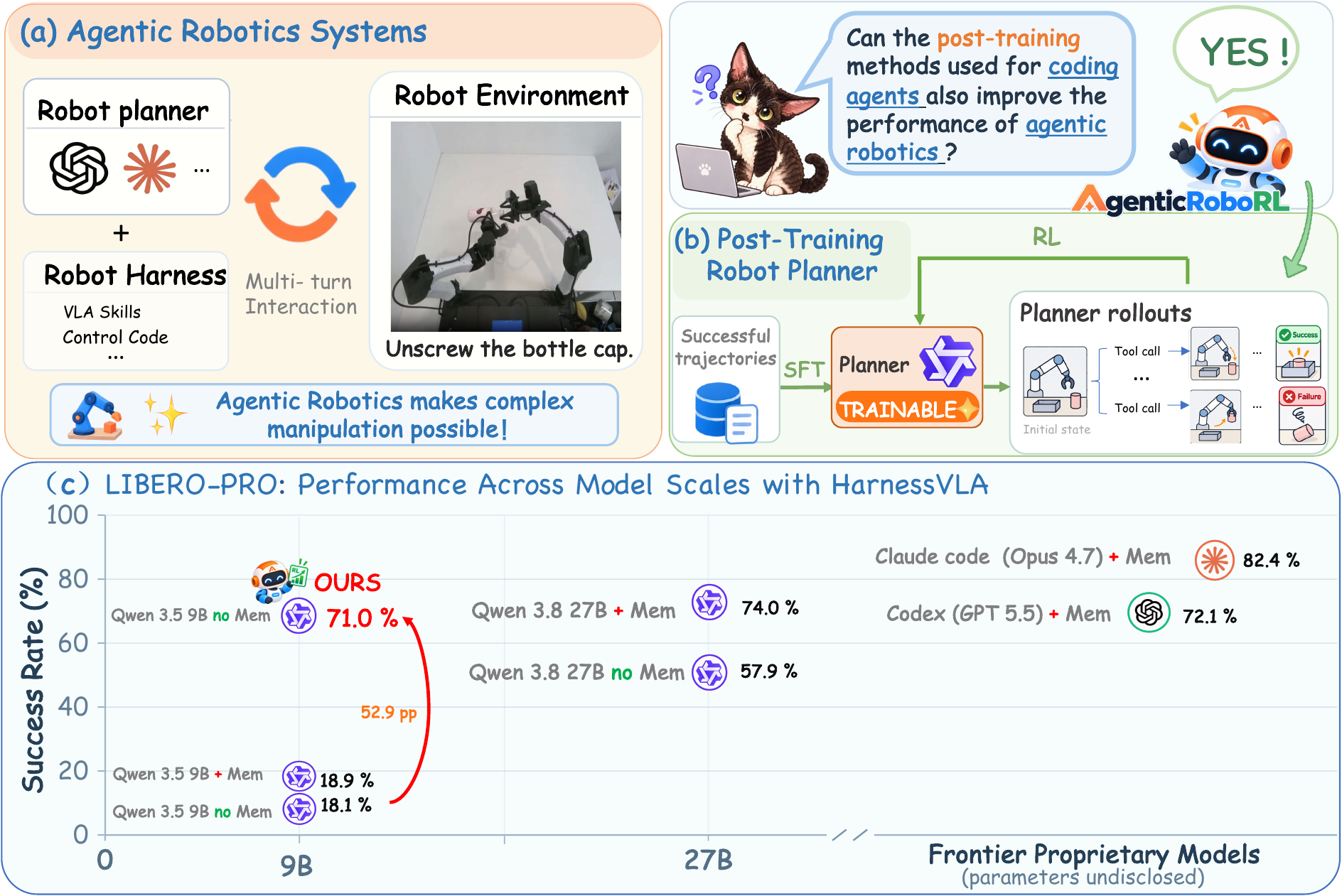}
\caption{\textbf{Overview of \ours.} (a) Agentic robotics systems combine planners with robot harnesses for multi-turn interaction with the environment. (b) We post-train planners through SFT on successful trajectories followed by multi-turn RL. (c) On LIBERO-PRO, post-training improves a 9B planner’s success rate from 18.1\% to 71.0\%, approaching the 72.1\% achieved by Codex (GPT-5.5)with memory. Mem denotes external memory.}
\label{fig:teaser}
\end{figure*}

\looseness=-1 Agentic robotics systems can perform complex manipulation tasks by combining a planner with a robot harness. The planner invokes VLA skills~\citep{harnessvla} or generates control code~\citep{codeaspolicies,capx}, observes the execution results, and decides what to do next over multiple rounds of interaction. Recent work has improved performance by refining robot harnesses and storing experience from task exploration in external memory~\citep{harnessvla,baton}, with coding agents such as Codex and Claude Code~\citep{codex,claudecode} serving as planners. However, these approaches primarily leverage existing planner capabilities and strengthen the support available to the planner. We explore a complementary direction: enhancing planners’ own capabilities to improve the performance of agentic robotics systems.

For coding agents, enhancing the model's own capabilities through post-training, including supervised fine-tuning and reinforcement learning, has proven highly effective~\citep{swegym,swesmith,skyrlagent}. These advances raise a natural question: \emph{\textbf{Can the post-training methods used for coding agents also improve the performance of \mbox{agentic robotics systems?}}}

Applying these methods to robot planners is complicated by the fact that robot harnesses differ in how they carry out a planner's response. In some harnesses the planner calls learned VLA skills, and a VLA policy executes each call. In others the planner writes control code over perception and control APIs, and a motion controller executes it. Harnesses also differ in what feedback they return to the planner and in how they decide whether the task has succeeded. Training planners across harnesses therefore requires a single pipeline that supports these different execution mechanisms.

To realize such a pipeline, we introduce \ours, a framework for post-training robot planners across different harnesses (Figure~\ref{fig:teaser}). The training recipe combines supervised fine-tuning (SFT) with multi-turn reinforcement learning (RL). SFT trains the planner on successful demonstrations, providing a starting point for RL. During RL, we train on recorded planner requests and responses, while harness-specific adapters associate these interactions with episode outcomes. This allows the same RL pipeline to support VLA skill invocation and control-code generation without changing their execution interfaces. In both stages, only the planner is optimized, while the underlying execution components remain fixed (Figure~\ref{fig:framework}).

We evaluate the pipeline across these two execution mechanisms using a 9B planner. With RPent~\citep{harnessvla}, a harness that invokes VLA skills, SFT improves success on LIBERO-PRO from 18.1\% to 65.1\%, and RL further raises it to 71.0\%, without external memory. The same checkpoint achieves 72.2\% on LIBERO-Plus~\citep{liberoplus} without additional training, compared with 19.2\% for the original model. With the CaP-X harness~\citep{capx}, post-training improves success from 2.7\% to 28.5\%. \textbf{These results establish post-training as an effective approach for improving agentic robotics systems.}

Our contributions are threefold:
\begin{itemize}\setlength{\itemsep}{0pt}\setlength{\parsep}{0pt}
    \item To our knowledge, \texttt{\textbf{AgenticRoboRL}} is the first general post-training framework to agentic robotics systems.
    \item \texttt{\textbf{AgenticRoboRL}} first fine-tunes the planner on successful demonstrations and then refines it with RL, updating only the planner while keeping execution components fixed. Lightweight harness-specific adapters link each recorded planner call to its episode outcome, so the harnesses do not need a shared low-level action representation.
    \item Strong empirical results and an open release. With \texttt{\textbf{AgenticRoboRL}}, a 9B planner improves from 18.1\% to 71.0\% on LIBERO-PRO without external memory. This nearly matches Codex (GPT-5.5) with external memory (72.1\%)~\citep{harnessvla}. With the CaP-X harness, it improves from 2.7\% to 28.5\%. We will release the \texttt{\textbf{AgenticRoboRL}} framework and trained planner checkpoints.
\end{itemize}

\section{Related Work}
\label{sec:related}

\paragraph{Robot harness agents.}
Language models can guide robot behavior by selecting learned skills or generating programs over perception and control APIs~\citep{saycan,codeaspolicies,progprompt,voxposer,robocodex}. In this organization, the planner chooses how to use the available capabilities, while the harness executes its decisions and returns observations. Execution feedback and reusable skill libraries support interaction over multiple turns~\citep{innermonologue,voyager}. Hierarchical systems place a language-level reasoning model above a learned low-level policy~\citep{rth,hirobot}. Recent harnesses expose these capabilities to coding agents: Harness VLA combines a frozen VLA with analytic primitives and task memory~\citep{harnessvla}, while CaP-X provides interfaces for generating executable control code~\citep{capx}. General-purpose coding agents have also been applied to manipulation without robot-specific training~\citep{aor,agentaspolicy}. Other systems improve perception, memory, safety, or skill libraries around the planner~\citep{roboharnessk1,baton,dynaharness,mtp,skillcomposer}. A concurrent training-free approach, RV-ICL, provides on-demand access to a hierarchy of demonstration-video keyframes and clips~\citep{rvicl}. \ours instead improves the planner through SFT and multi-turn RL while reusing the harness's \mbox{execution components.}

\paragraph{Training planners for embodied agents.}
RL has been used to improve language and vision-language planners in embodied environments~\citep{embodiedplannerr1,era,robogptr1,reinforcedreasoning,dreamplan}. CaP-RL applies RL to a code model that generates a complete control program in a single turn~\citep{capx}. In contrast, \ours trains planners that interact with the harness over multiple turns, and it applies to harnesses based on VLA skills as well as to harnesses based on control code. Closer to our supervised stage, Robo-Harness K1 and World Action Agent train 9B students on teacher trajectories collected within their respective harnesses~\citep{roboharnessk1,worldactionagent}. These methods use supervised training to transfer the capabilities of stronger agents. REAL combines SFT with online RL for mobile manipulation, using a rule-based planner for supervision and a shaped reward~\citep{real}. \ours instead uses teacher-agent trajectories followed by multi-turn RL with task success as the sole reward. It adapts teacher supervision when the teacher and student have different prompts, tools, or access to memory.\looseness=-1

\paragraph{Post-training of tool-using agents.}
Coding-agent training commonly combines supervision from successful teacher trajectories~\citep{swegym,swesmith,r2egym} with multi-turn RL on verifiable outcomes~\citep{skyrlagent}. Related approaches extend to search and computer-use agents~\citep{searchr1,opensearchvl,uitars2,computerrl}. Frameworks including Agent Lightning and LEGO-RL separate agent execution from model optimization and construct training data from recorded interactions~\citep{agentlightning,agentlightningv1,legorl}. \ours builds on this approach for robot harnesses: a proxy records planner calls, and harness-specific adapters associate them with episode outcomes. This interface accommodates different execution mechanisms without requiring a common low-level action representation.

\section{\texorpdfstring{\ours}{AgenticRoboRL} Framework}
\label{sec:method}

\begin{figure*}[t]
\centering
\includegraphics[width=\textwidth]{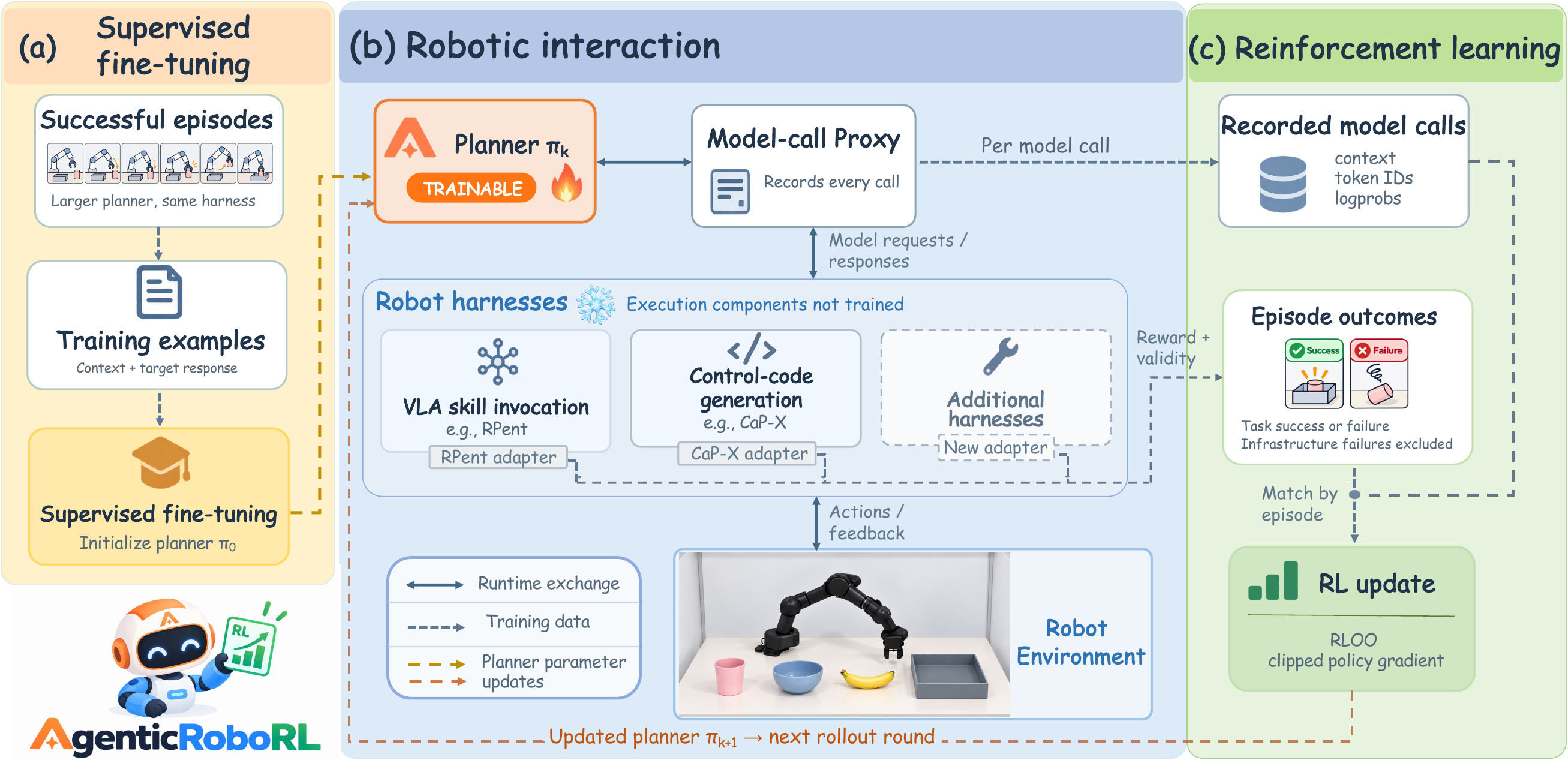}
\caption{\textbf{\ours training pipeline.} (a) Successful trajectories are converted into training examples for SFT. (b) The planner interacts with the environment through a robot harness, while a proxy records its model calls. (c) Recorded calls are matched with episode outcomes for RL updates, excluding episodes invalidated by infrastructure failures. Harness-specific adapters support a common training pipeline across different harnesses. Only the planner is trained; execution components remain fixed.}
\label{fig:framework}
\end{figure*}

\subsection{Overview and Problem Setup}

\ours post-trains the planner of a robot agent while reusing its harness's execution components. The planner produces harness-native responses, including structured tool calls in RPent, the harness released with Harness VLA~\citep{harnessvla}, or executable Python code in CaP-X~\citep{capx}. The harness executes these commands and returns observations and execution feedback for subsequent decisions. We update the planner, not the perception modules, motion planners, or VLA executors it invokes. Figure~\ref{fig:framework} summarizes the training pipeline: teacher supervision initializes the planner, and multi-turn interaction supplies task outcomes for reinforcement learning (RL).

Let $\xi_i=(x_i,s_i^0)$ specify a task and its initial-state configuration. At turn $t$, the harness $\mathcal H$ constructs the actual model context $c_{i,t}$ from the task, observations, and preceding responses:
\begin{equation}
\begin{aligned}
c_{i,t} &= \Phi_{\mathcal H}(x_i,o_{i,\leq t},y_{i,<t}),\\
y_{i,t} &\sim \pi_\theta(\cdot\mid c_{i,t}).
\end{aligned}
\label{ar:context}
\end{equation}
Here $\Phi_{\mathcal H}$ includes prompt construction, tool specifications, and history processing. A response $y_{i,t}$ is a complete model-generated token sequence, potentially containing reasoning, text, tool calls, or control code; it is not a single low-level robot action. A trajectory $\tau_i$ contains the successive model requests and responses together with their execution outcomes. Our objective is
\begin{equation}
\begin{aligned}
\max_\theta\ J(\theta)
&= \mathbb E_{\substack{\xi\sim\mathcal D,\\
\tau\sim P_{\theta,\mathcal H}(\cdot\mid\xi)}}[R(\tau)],\\
R(\tau)&=\mathbf 1[\text{task success}].
\end{aligned}
\label{ar:objective}
\end{equation}
$\mathcal D$ denotes the training task/initial-state distribution. This objective describes the terminal-success reward used in our main RL configuration, without intermediate rewards or a length penalty.

\subsection{Training Infrastructure}

\paragraph{Recording the actual model interaction.}
Because VLA skill invocations and generated control programs differ in representation and execution granularity, we place the shared RL interface at the model-call boundary rather than the low-level action boundary. During RL, the training unit is a recorded planner request, not a simulator step or a reconstructed full transcript. A recording proxy captures each request's context, generated token IDs, and rollout log-probabilities at the model-serving boundary, following recent agent training frameworks~\citep{agentlightning,legorl,polar}. Images and tool specifications are retained where applicable. This matters because a harness may prune images, omit previous reasoning, or otherwise transform its history between requests. The trainer checks its rendered prompt against the recorded prompt tokens; mismatched samples receive zero optimization weight. Training therefore uses the inputs associated with each sampled response rather than assuming that a final transcript reproduces them.

\paragraph{Connecting interactions to outcomes.}
A harness-specific adapter reports, for every episode, five fields: the task, the initial-state identifier, the attempt index, the reward, and a validity flag. The first two define the group used for within-group comparison; the validity flag separates task failures, which remain reward-zero episodes, from infrastructure failures, which are excluded from optimization. Recorded model calls are then matched to their episode, and the pipeline checks record completeness and policy version before any sample is built. The adapter is plain conversion code with no learnable parameters: it provides a common training representation without imposing a common robot execution interface, and it does not alter the harness's tools, prompts, or control loop. Appendix~\ref{app:adapters} lists the fields for our two harnesses.

\paragraph{Policy-versioned rounds.}
Each round collects episodes under fixed planner weights, constructs training samples from the recorded interactions, and updates the planner before the next collection round. Multiple attempts share a task and initial-state configuration for within-group comparison. Collection, outcome processing, and optimization are separate stages; the planner is not updated during an episode.\looseness=-1

\subsection{Planner Post-Training}

\subsubsection{Supervised Fine-Tuning}
\label{sec:method-sft}

Teacher trajectories provide an initial policy for using the harness (Figure~\ref{fig:framework}(a)). A training example pairs a teacher response with the context the \emph{student} would receive at that turn. This context can differ from what the teacher saw, because the two planners may run under different harness configurations or chat templates. How the student context is obtained depends on the harness.

\textit{RPent.} The teacher runs with a memory module that the student does not have. We therefore reconstruct each turn's student context offline from the recorded episode, removing memory-access calls and their results. A response is flagged as dependent on memory if it contains keywords that refer to the memory, repeats eight consecutive words of the memory text, or contains a number that appears only in a removed memory result. Teacher reasoning and text that refer to memory are rewritten and checked by a judge, while tool-call arguments are kept unchanged. Only rewrites that the judge labels OK become training targets. After this processing, each turn is used as a training target, kept only as history, or dropped. Section~\ref{sec:exp-supervision} compares this recipe with excluding memory-dependent responses from supervision.

\textit{CaP-X.} Teacher and student share the same harness configuration, so no memory-related context reconstruction is needed. We take the recorded model calls of successful episodes, keep those that ended normally, and re-render each prompt with the student's chat template. The teacher's recorded output tokens are used directly as the target.

For accepted target $s$, let $c_s$ be the actual student context and $y_s$ the teacher response after any prescribed processing. With $L_s$ supervised tokens, the SFT loss is
\begin{equation}
\mathcal L_{\mathrm{SFT}}=
-\frac{\displaystyle\smashoperator[l]{\sum_{s\in\mathcal B_{\mathrm{SFT}}}}\sum_{j=1}^{L_s}
\log\pi_\theta(y_{s,j}\mid
c_s,y_{s,<j})}
{\displaystyle\sum_{s\in\mathcal B_{\mathrm{SFT}}}L_s}.
\label{ar:sft}
\end{equation}
Only accepted response tokens, including reasoning, text, serialized tool calls or code, and the end token, receive supervision. Context tokens are masked. The denominator counts supervised tokens across the optimizer batch, rather than averaging independently normalized response losses.

\subsubsection{Reinforcement Learning through Interaction}

The initialized planner collects multi-turn trajectories through the harness (Figure~\ref{fig:framework}(b)). For valid attempts $\mathcal V_g$ sharing a task and initial-state configuration, let $G_g=|\mathcal V_g|\geq2$. The terminal reward $R_i$ yields the leave-one-out advantage~\citep{rloo}
\begin{equation}
A_i=R_i-\frac{1}{G_g-1}
\sum_{k\in\mathcal V_g\setminus\{i\}}R_k.
\label{ar:rloo}
\end{equation}
We use neither a learned critic nor reward-standard-deviation normalization. Every selected response in episode $i$ receives $A_i$; groups with identical rewards contribute no policy gradient.

To limit training cost, we select at most 16 eligible responses per episode. Eligibility follows the configured record-validity, token-length, and post-success filters. Let $\mathcal S_i$ be the selected turns and $K_i=|\mathcal S_i|$. For a selected response $s$, let $i(s)$ identify its episode and $u_s=1/K_{i(s)}$. The update in Figure~\ref{fig:framework}(c) uses
\begin{equation}
\mathcal L_{\mathrm{RL}}=
\frac{\displaystyle\sum_{s\in\mathcal B_{\mathrm{RL}}}
\frac{u_s}{L_s}\sum_{j=1}^{L_s}
w_{sj}\ell_{\mathrm{DC}}(\rho_{sj},A_{i(s)})}
{\displaystyle\sum_{s\in\mathcal B_{\mathrm{RL}}}u_s}.
\label{ar:rl}
\end{equation}
Here $\mathcal B_{\mathrm{RL}}$ contains the non-padding responses with nonzero effective training weight in one optimizer step. $L_s$ counts response tokens, $\rho$ is the current-to-old trainer probability ratio, $w$ mitigates probability mismatch between the rollout and training backends, and $\ell_{\mathrm{DC}}$ is the dual-clipped PPO token loss~\citep{ppo,dualclipppo}~\mbox{(Appendix~\ref{app:policy-details}).}

% Section 4, version 2 (2026-10-10). English not yet polished.
% Needs: \usepackage{booktabs,graphicx,amsmath}.
% Citation keys used here: harnessvla, liberopro, liberoplus, capx, mcnemar1947, and the keys cited in the tables.
% TODO: Table 3 numbers may change (perception re-evaluation running; RL version 18 has one evaluation = 28.8).
% TODO: 64.6 in Table 4 and the 11 leaderboard rows of Table 2 still need a check against their sources.
\section{Experiments}
\label{sec:experiments}

\subsection{Experimental Setup}
\label{sec:exp-setup}

\paragraph{Robot environments and harnesses.}
Our main experiments use RPent, the VLA-calling harness released with Harness VLA~\citep{harnessvla}, and evaluate planners in it on LIBERO-PRO~\citep{liberopro}. We additionally evaluate the base, SFT, and RL checkpoints on the 10{,}030 configurations of LIBERO-Plus~\citep{liberoplus} to measure robustness without benchmark-specific training. To demonstrate the generality of the framework, we also attach it to a second harness, CaP-X~\citep{capx}; its setup and results are given in Section~\ref{sec:exp-capx}.

\paragraph{Models and comparisons.}
We use Qwen3.5-9B~\citep{qwen35} as the student planner and compare the original checkpoint with its SFT and RL counterparts. Qwen3.8-27B~\citep{qwen38} provides teacher trajectories and serves as a larger-model performance reference. The execution components, including~the~VLA,~\mbox{remain}~frozen.

\paragraph{Memory.}
Harness VLA provides task-specific memory of successful action sequences and global memory of rules and failure cases shared across tasks. Our 27B teacher uses the released memory files. For RPent, the 9B student is supervised using trajectories from this teacher. During RL rollouts and evaluation, the student does not access external memory.\looseness=-1

\paragraph{Data split.}
Each LIBERO~\citep{libero} task has 50 fixed initial states. Our evaluation follows the task-specific experience setting of Harness VLA~\citep{harnessvla}, where the agent explores each task on initial state 0 to construct memory and is evaluated on initial states 1--10. We retain the same task set and evaluation initial states, while extending this setting with planner post-training. Teacher demonstrations and RL rollouts are collected on states 11--44, which do not overlap with the evaluation initial states. States 45--49 are used only as a validation set for checkpoint and task selection. LIBERO-Plus was run after training settings were fixed. It uses the same underlying LIBERO tasks and adds perturbations that the planner \mbox{did not see during training.}

\paragraph{Training.}
For supervised fine-tuning we use AdamW with a learning rate of $10^{-5}$ and a batch of 64 target turns, and we train for one epoch, which is 626 optimizer steps with RPent and 37 with CaP-X. RL uses a learning rate of $10^{-6}$ and a batch of 128 selected responses. Each RL round collects six attempts for each of 64 groups, where a group is one task with one initial state, and then updates the planner. This gives 384 episodes per round. Training up to the reported RPent checkpoint collects 2{,}304 episodes, selects 26{,}052 responses, and takes 206 optimizer steps. With RPent the reported RL checkpoint is selected by its success rate on the validation states 45--49. The SFT checkpoint at step 563 is selected among ten checkpoints on the same validation states. With CaP-X the reported planner is trained for 15 rounds. With RPent an episode ends after at most 100 planner turns or 7{,}200 seconds. Appendix~\ref{app:training} gives the remaining settings.

\subsection{Results on LIBERO-PRO}
\label{sec:exp-liberopro}

LIBERO-PRO evaluates whether policies adapt to changes in the scene and task instruction rather than relying on memorized motion patterns. Following Harness VLA, we evaluate two perturbation types across the four LIBERO suites: position swap (S), which exchanges object positions, and task perturbation (T), which changes the instruction to specify a different target. This yields eight evaluation suites, each containing ten tasks evaluated on ten initial states, for a total of 800 episodes.

\colorlet{oursblue}{blue!15}
\definecolor{gaingreen}{HTML}{008C3C}
\newcommand{\gain}[1]{{\hspace{1.5pt}\fontsize{7pt}{8pt}\selectfont\bfseries\textcolor{gaingreen}{+#1}}}
\begin{table*}[t]
\centering
\small
\setlength{\tabcolsep}{4pt}
\resizebox{\textwidth}{!}{%
\begin{tabular}{@{}c@{\hspace{6pt}}lc*{9}{r@{}l}@{}}
\toprule
 & \textbf{Method} & \textbf{Mem.} & \multicolumn{1}{c@{}}{\textbf{Spat-T}} & & \multicolumn{1}{c@{}}{\textbf{Spat-S}} & & \multicolumn{1}{c@{}}{\textbf{Obj-T}} & & \multicolumn{1}{c@{}}{\textbf{Obj-S}} & & \multicolumn{1}{c@{}}{\textbf{Goal-T}} & & \multicolumn{1}{c@{}}{\textbf{Goal-S}} & & \multicolumn{1}{c@{}}{\textbf{L10-T}} & & \multicolumn{1}{c@{}}{\textbf{L10-S}} & & \multicolumn{1}{c@{}}{\textbf{Overall}} & \\
\midrule
 & OpenVLA~\citep{openvla}$^\dagger$ & -- & 0.0 & & 0.0 & & 0.0 & & 0.0 & & 0.0 & & 0.0 & & 0.0 & & 0.0 & & 0.0 & \\
 & $\pi_0$~\citep{pi0}$^\dagger$ & -- & 0.0 & & 0.0 & & 0.0 & & 2.0 & & 0.0 & & 0.0 & & 0.0 & & 0.0 & & 0.3 & \\
 & $\pi_{0.5}$~\citep{pi05}$^\dagger$ & -- & 1.0 & & 20.0 & & 1.0 & & 17.0 & & 2.0 & & 38.0 & & 1.0 & & 8.0 & & 11.0 & \\
 & MolmoAct~\citep{molmoact}$^\dagger$ & -- & 0.0 & & 0.0 & & 0.0 & & 6.0 & & 0.0 & & 0.0 & & 6.0 & & 0.0 & & 1.5 & \\
 & NORA~\citep{nora}$^\dagger$ & -- & 0.0 & & 0.0 & & 0.0 & & 0.0 & & 0.0 & & 0.0 & & 0.0 & & 0.0 & & 0.0 & \\
 & X-VLA~\citep{xvla}$^\dagger$ & -- & 0.0 & & 0.0 & & 8.0 & & 2.0 & & 9.0 & & 1.0 & & 10.0 & & 0.0 & & 3.8 & \\
 & AtomVLA~\citep{atomvla}$^\dagger$ & -- & 1.0 & & 16.0 & & 0.0 & & 10.0 & & 11.0 & & 2.0 & & 9.0 & & 1.0 & & 6.3 & \\
\multirow{-8}{*}{\rotatebox[origin=c]{90}{VLA}} & $\pi_{\mathrm{RLinf}}$ alone~\citep{rlinf}$^\dagger$ & -- & 42.0 & & 59.0 & & 71.0 & & 78.0 & & 45.0 & & 42.0 & & 49.0 & & 14.0 & & 50.0 & \\
\midrule
 & CaP-X~\citep{capx}$^\dagger$ & -- & 14.0 & & 12.0 & & 18.0 & & 22.0 & & 17.0 & & 26.0 & & \multicolumn{1}{c@{}}{/} & & \multicolumn{1}{c@{}}{/} & & 18.2 & \\
 & RATS~\citep{rats}$^\dagger$ & -- & 31.0 & & 29.0 & & 63.0 & & 61.0 & & 36.0 & & 43.0 & & \multicolumn{1}{c@{}}{/} & & \multicolumn{1}{c@{}}{/} & & 43.8 & \\
 & Harness VLA, Codex (GPT-5.5)$^\dagger$ & yes & 81.0 & & 69.0 & & 94.0 & & 91.0 & & 75.0 & & 66.0 & & 52.0 & & 49.0 & & 72.1 & \\
 & Harness VLA, Claude Code (Opus 4.7)$^\dagger$ & yes & 94.0 & & 80.0 & & 88.0 & & 90.0 & & \textbf{87.0} & & \textbf{87.0} & & \textbf{71.0} & & \textbf{62.0} & & \textbf{82.4} & \\
\cmidrule[\lightrulewidth](l{0pt}){2-21}
 & RPent + 27B (teacher) & yes & 89.0 & & 74.0 & & 93.0 & & 85.0 & & 67.0 & & 66.0 & & 66.0 & & 52.0 & & 74.0 & \\
 & RPent + 27B & no & 71.0 & & 68.0 & & 85.0 & & 85.0 & & 51.0 & & 42.0 & & 33.0 & & 28.0 & & 57.9 & \\
 & RPent + 9B & yes & 13.0 & & 21.0 & & 26.0 & & 25.0 & & 26.0 & & 23.0 & & 13.0 & & 4.0 & & 18.9 & \\
\arrayrulecolor{black!35}\cmidrule[\lightrulewidth](l{0pt}){2-21}\arrayrulecolor{black}
 & RPent + 9B & no & 22.0 & & 17.0 & & 19.0 & & 15.0 & & 24.0 & & 35.0 & & 9.0 & & 4.0 & & 18.1 & \\
 & RPent + 9B, SFT (ours) & no & 86.0 & \gain{64.0} & 82.0 & \gain{65.0} & 94.0 & \gain{75.0} & 85.0 & \gain{70.0} & 54.0 & \gain{30.0} & 62.0 & \gain{27.0} & 32.0 & \gain{23.0} & 26.0 & \gain{22.0} & 65.1 & \gain{47.0} \\
\rowcolor{oursblue}[0pt][2\tabcolsep]
\cellcolor{white}\multirow{-10}{*}{\rotatebox[origin=c]{90}{Agent}} & RPent + 9B, SFT + RL (ours) & no & \textbf{96.0} & \gain{74.0} & \textbf{86.0} & \gain{69.0} & \textbf{96.0} & \gain{77.0} & \textbf{95.0} & \gain{80.0} & 54.0 & \gain{30.0} & 61.0 & \gain{26.0} & 41.0 & \gain{32.0} & 39.0 & \gain{35.0} & 71.0 & \multicolumn{1}{@{}>{\columncolor{oursblue}[0pt][0pt]}l@{}}{\gain{52.9}} \\
\bottomrule
\end{tabular}%
}
\caption{\textbf{Success rate (\%) on LIBERO-PRO.} Each cell aggregates 100 episodes (10 tasks $\times$ 10 initial states). T: task perturbation; S: position swap. Rows marked $^\dagger$ are taken from Harness VLA~\citep{harnessvla}; ``/'' means not reported. Mem.: the planner retrieves memory at test time. RPent is the harness released with Harness VLA: the Harness VLA rows and the RPent rows use the same harness and the same frozen VLA ($\pi_{\mathrm{RLinf}}$), and differ only in the planner. The row shaded in blue is our final planner. The best entry in each column is in bold.}
\label{tab:liberopro}
\end{table*}

Table~\ref{tab:liberopro} reports the success rates on LIBERO-PRO. SFT improves the 9B planner from 18.1\% to 65.1\%, and subsequent RL raises its success rate to 71.0\%. SFT accounts for most of the improvement, while RL adds another 5.9 percentage points. The post-trained 9B planner outperforms the original 27B planner without memory (57.9\%) and comes within 3.0 percentage points of the 27B teacher with memory (74.0\%). It also outperforms the standalone VLA, which achieves 50.0\% overall, with higher success rates in seven of the eight suites. On the 800 paired episodes, the RL planner solves 121 episodes that the SFT planner fails and fails 74 that it solves (exact sign test, $p = 0.001$).

The student can learn task knowledge from the trajectories of the teacher with memory, although it does not retrieve external memory at test time. Comparisons with planners that use memory provide system-level references rather than isolating the contribution of supervision from~a~teacher~with~memory.

% TODO figure: LIBERO-PRO success per RL round. Evaluated so far (one 800-episode evaluation each):
% round 3: 68.2, round 6: 71.0, round 7: 69.2, round 9: 66.6. Rounds 1, 2, 4, 5, 8 not evaluated.

\subsection{Robustness on LIBERO-Plus}
\label{sec:exp-liberoplus}

To evaluate robustness to perturbations that are absent from training, we evaluate the same checkpoints used on LIBERO-PRO on LIBERO-Plus without additional training. LIBERO-Plus introduces seven categories of perturbations across the four LIBERO suites: object layout, camera viewpoint, robot initial state, language instruction, lighting, background texture, and sensor noise. Following the official protocol, we run one episode for each of the 10{,}030 evaluation instances.

As shown in Table~\ref{tab:liberoplus}, the original 9B planner achieves 19.2\% success, compared with 66.7\% after SFT and 72.2\% after RL. Both training stages improve performance in every perturbation category, with RL adding between 2.2 percentage points on object layout and 7.0 points on lighting. Comparing outcomes on the same evaluation instances, RL succeeds on 1{,}547 instances where SFT fails, but fails on 999 instances where SFT succeeds, yielding a net gain of 548 successful episodes.\looseness=-1

\begin{table*}[t]
\centering
\small
\resizebox{\textwidth}{!}{%
\begin{tabular}{@{}c@{\hspace{6pt}}l*{8}{r@{}l}@{}}
\toprule
 & \textbf{Method} & \multicolumn{1}{c@{}}{\textbf{Camera}} & & \multicolumn{1}{c@{}}{\textbf{Robot}} & & \multicolumn{1}{c@{}}{\textbf{Language}} & & \multicolumn{1}{c@{}}{\textbf{Light}} & & \multicolumn{1}{c@{}}{\textbf{Background}} & & \multicolumn{1}{c@{}}{\textbf{Noise}} & & \multicolumn{1}{c@{}}{\textbf{Layout}} & & \multicolumn{1}{c@{}}{\textbf{Total}} & \\
\midrule
 & OpenVLA~\citep{openvla}$^\dagger$ & 0.8 &  & 3.5 &  & 23.0 &  & 8.1 &  & 34.8 &  & 15.2 &  & 28.5 &  & 15.6 &  \\
 & OpenVLA-OFT~\citep{openvlaoft}$^\dagger$ & 56.4 &  & 31.9 &  & 79.5 &  & 88.7 &  & 93.3 &  & 75.8 &  & \textbf{74.2} &  & 69.6 &  \\
 & OpenVLA-OFT\_w$^\dagger$ & 10.4 &  & 38.7 &  & 70.5 &  & 76.8 &  & \textbf{93.6} &  & 49.9 &  & 69.9 &  & 55.8 &  \\
 & NORA~\citep{nora}$^\dagger$ & 2.2 &  & 37.0 &  & 65.1 &  & 45.7 &  & 58.6 &  & 12.8 &  & 62.1 &  & 39.0 &  \\
 & WorldVLA~\citep{worldvla}$^\dagger$ & 0.1 &  & 27.9 &  & 41.6 &  & 43.7 &  & 17.1 &  & 10.9 &  & 38.0 &  & 25.0 &  \\
 & UniVLA~\citep{univla}$^\dagger$ & 1.8 &  & 46.2 &  & 69.6 &  & 69.0 &  & 81.0 &  & 21.2 &  & 31.9 &  & 43.9 &  \\
 & $\pi_0$~\citep{pi0}$^\dagger$ & 13.8 &  & 6.0 &  & 58.8 &  & 85.0 &  & 81.4 &  & \textbf{79.0} &  & 68.9 &  & 53.6 &  \\
 & $\pi_0$-Fast~\citep{pi0fast}$^\dagger$ & 65.1 &  & 21.6 &  & 61.0 &  & 73.2 &  & 73.2 &  & 74.4 &  & 68.8 &  & 61.6 &  \\
 & RIPT-VLA~\citep{riptvla}$^\dagger$ & 55.2 &  & 31.2 &  & 77.6 &  & 88.4 &  & 91.6 &  & 73.5 &  & \textbf{74.2} &  & 68.4 &  \\
 & OpenVLA-OFT\_m$^\dagger$ & 55.6 &  & 21.7 &  & \textbf{81.0} &  & \textbf{92.7} &  & 91.0 &  & 78.6 &  & 68.7 &  & 67.9 &  \\
\multirow{-11}{*}{\rotatebox[origin=c]{90}{VLA}} & OpenVLA-OFT+ (trained on LIBERO-Plus)$^\dagger$ & 92.8 &  & 30.3 &  & 85.8 &  & 94.9 &  & 93.9 &  & 89.3 &  & 77.6 &  & 79.6 &  \\
\midrule
 & RPent + 9B & 18.0 &  & 20.5 &  & 19.4 &  & 21.0 &  & 17.3 &  & 17.2 &  & 21.2 &  & 19.2 &  \\
 & RPent + 9B, SFT (ours) & 59.2 & \gain{41.2} & 70.6 & \gain{50.1} & 67.9 & \gain{48.5} & 68.9 & \gain{47.9} & 66.8 & \gain{49.5} & 66.2 & \gain{49.0} & 68.4 & \gain{47.2} & 66.7 & \gain{47.5} \\
\rowcolor{oursblue}[0pt][2\tabcolsep]
\cellcolor{white}\multirow{-3}{*}{\rotatebox[origin=c]{90}{Agent}} & RPent + 9B, SFT + RL (ours) & \textbf{65.8} & \gain{47.8} & \textbf{75.2} & \gain{54.7} & 73.9 & \gain{54.5} & 75.9 & \gain{54.9} & 72.0 & \gain{54.7} & 73.0 & \gain{55.8} & 70.6 & \gain{49.4} & \textbf{72.2} & \multicolumn{1}{@{}>{\columncolor{oursblue}[0pt][0pt]}l@{}}{\gain{53.0}} \\
\bottomrule
\end{tabular}%
}
\caption{\textbf{Success rate (\%) on LIBERO-Plus by perturbation category.} Rows marked $^\dagger$ are from the LIBERO-Plus leaderboard~\citep{liberoplus} and are single VLA policies; our rows are a planner with the same frozen VLA as in Table~\ref{tab:liberopro} ($\pi_{\mathrm{RLinf}}$). The green numbers give the gain over the untrained RPent + 9B row. The row shaded in blue is our final planner. Among methods not trained on LIBERO-Plus, the best entry in each column is in bold.}
\label{tab:liberoplus}
\end{table*}

Among the methods compared in Table~\ref{tab:liberoplus} that do not use LIBERO-Plus for training, our RL planner achieves the highest overall success rate. Its largest margin over the best compared baseline is on robot initial-state perturbations, where it achieves 75.2\% versus 46.2\%. These results show that the gains from post-training remain under the perturbations of LIBERO-Plus.

\subsection{Generality across Harnesses}
\label{sec:exp-capx}

\paragraph{Setup.}
To evaluate whether \ours applies beyond VLA skill invocation, we use it to train a planner in CaP-X~\citep{capx}, where the planner generates executable control code. We integrate the released CaP-X harness through the model-call proxy and a harness-specific adapter, reusing the training procedure described in Section~\ref{sec:method} and Figure~\ref{fig:framework}. We compare the original, SFT, and RL checkpoints of the 9B planner~within~this~harness.

Following the task selection in CaP-X, we use the Object, Goal, and Spatial suites of LIBERO-PRO under position and task perturbations, comprising 60 tasks. Following the setting of CaP-RL, the RL experiment in the CaP-X paper~\citep{capx}, we use privileged state. The teacher and the original, SFT, and RL planners all use this setting during trajectory collection and evaluation, so the comparisons in Table~\ref{tab:capx} hold the execution interface fixed. The 27B planner provides teacher trajectories, and neither the teacher nor the student uses external task memory in this setting. We train the 9B planner through SFT followed by RL, using trajectories collected in CaP-X rather than transferring the RPent-trained checkpoints. We retain the same initial-state split: states 11--44 for teacher trajectories and RL rollouts, and states 1--10 for evaluation. Each cell of Table~\ref{tab:capx} therefore aggregates 100 episodes (10 tasks $\times$ 10 initial states).

\paragraph{Results.}
Table~\ref{tab:capx} shows that SFT improves the 9B planner's success rate from 2.7\% to 18.7\%, and RL further raises it to 28.5\% (our final planner, shaded in blue), exceeding the 27B teacher's 20.3\%. The teacher data covers only 26 of the 60 tasks, namely 19 of the 20 Object tasks, five of the 20 Goal tasks, and two of the 20 Spatial tasks, and yields 979 episodes for SFT. RL requires no teacher data and raises the student above the teacher. Improvements are concentrated in the two Object suites, where RL raises success rates from 39.0--49.0\% to 67.0--72.0\%. Performance remains low on Goal and Spatial, where the teacher rarely succeeds and therefore provides little SFT data. The low absolute numbers thus reflect the difficulty of this harness on LIBERO-PRO rather than a failure of post-training. These results demonstrate that the same post-training procedure can improve planners that generate control code as well as those that invoke VLA skills.

\begin{table}[t]
\centering
\small
\renewcommand{\arraystretch}{1.25}
\setlength{\tabcolsep}{3pt}
\resizebox{\columnwidth}{!}{%
\begin{tabular}{@{}lccccccr@{}l@{}}
\toprule
 & \multicolumn{2}{c}{\textbf{Object}} & \multicolumn{2}{c}{\textbf{Goal}} & \multicolumn{2}{c}{\textbf{Spatial}} & & \\
\cmidrule(lr){2-3}\cmidrule(lr){4-5}\cmidrule(lr){6-7}
\textbf{Method} & \textbf{Pos} & \textbf{Task} & \textbf{Pos} & \textbf{Task} & \textbf{Pos} & \textbf{Task} & \multicolumn{1}{c@{}}{\textbf{Avg.}} & \\
\midrule
27B (teacher) & 41.0 & 53.0 & \textbf{12.0} & 10.0 & 6.0 & 0.0 & 20.3 &  \\
\arrayrulecolor{black!35}\midrule\arrayrulecolor{black}
9B & 5.0 & 5.0 & 3.0 & 1.0 & 1.0 & \textbf{1.0} & 2.7 &  \\
9B, SFT (ours) & 49.0 & 39.0 & 10.0 & 4.0 & \textbf{10.0} & 0.0 & 18.7 & \gain{16.0} \\
\rowcolor{oursblue}[0pt][2\tabcolsep]
9B, SFT + RL (ours) & \textbf{67.0} & \textbf{72.0} & 10.0 & \textbf{16.0} & 6.0 & 0.0 & \textbf{28.5} & \multicolumn{1}{@{}>{\columncolor{oursblue}[0pt][0pt]}l@{}}{\gain{25.8}} \\
\bottomrule
\end{tabular}%
}
\caption{\textbf{Success rate (\%) with the CaP-X harness on LIBERO-PRO.} Pos: position swap; Task: task perturbation. Green: gain over the 9B row. Bold: best per column.}
\label{tab:capx}
\end{table}

\subsection{Ablation Study}
\label{sec:exp-supervision}

\begin{figure}[t]
\centering
\IfFileExists{figures/fig_sft_curve.pdf}{\includegraphics[width=\columnwidth]{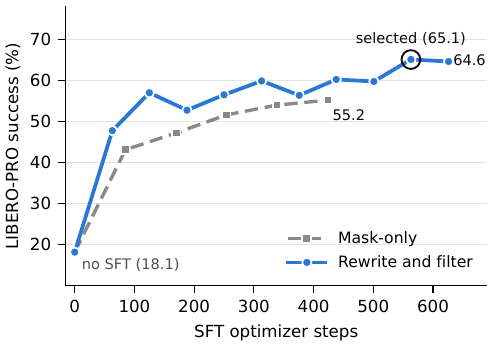}}{\fbox{Upload figures/fig\_sft\_curve.pdf}}
\vspace{-3pt}
\setlength{\abovecaptionskip}{4pt}
\caption{\textbf{Success on LIBERO-PRO during supervised fine-tuning for the two supervision recipes.} Each point is one evaluation on the 800 episodes. The circle marks the checkpoint used as the starting point for RL.}
\label{fig:sft-curve}

\vspace{3pt}

{\footnotesize\renewcommand{\arraystretch}{1.45}
\begin{tabular}{@{}lccc@{}}
\toprule
\textbf{Supervision recipe} & \textbf{Targets} & \textbf{Steps} & \textbf{Success} \\
\midrule
No SFT & -- & -- & 18.1 \\
Mask-only & 27{,}110 & 424 & 55.2 \\
Rewrite and filter & 40{,}047 & 438 & 60.2 \\
Rewrite and filter & 40{,}047 & 626 & \textbf{64.6} \\
\bottomrule
\end{tabular}}
\captionof{table}{\textbf{Ablation of the SFT supervision recipe on LIBERO-PRO (success, \%).} Table~\ref{tab:liberopro} reports the selected step-563 checkpoint of rewrite-and-filter (65.1\%).}
\label{tab:supervision}
\vspace{3pt}
\end{figure}

We compare two ways of handling teacher responses that depend on memory unavailable to the student, using the RPent harness on LIBERO-PRO. The first excludes these responses from supervision (mask-only). The second rewrites them and retains those accepted by a judge as training targets (rewrite-and-filter), following the procedure described in Section~\ref{sec:method-sft}. The resulting datasets contain 27{,}110 and 40{,}047 target turns, respectively. As shown in Table~\ref{tab:supervision}, after one epoch of training, mask-only improves the original 9B planner's success rate from 18.1\% to 55.2\%, while rewrite-and-filter reaches 64.6\%.

Because the datasets differ in size, one epoch corresponds to 424 optimizer steps for mask-only and 626 for rewrite-and-filter. We therefore also examine their learning curves (Figure~\ref{fig:sft-curve}) to compare performance at similar update counts. Rewrite-and-filter achieves 60.2\% at step 438, compared with 55.2\% for mask-only at step 424. The advantage is thus also observed at comparable update counts, rather than only at the end of the longer training run. However, the two recipes differ in both response content and supervision coverage, so this comparison does not isolate the contribution of rewriting alone.

\section{Conclusion}

We introduced \ours to investigate whether post-training methods used for coding agents can improve how robot planners use existing harnesses. The framework combines SFT on successful demonstration trajectories with multi-turn RL, connecting recorded model calls to episode outcomes while reusing the harnesses’ execution components. Experiments with a 9B planner demonstrate improvements in both VLA skill invocation and control-code generation. On LIBERO-PRO, success increases from 18.1\% to 71.0\%, and the same trained planner achieves 72.2\% on LIBERO-Plus without additional training. These results support post-training the planner as an effective way to improve robot-harness agents without retraining their underlying controllers. We will release the framework, teacher trajectories, and trained planner checkpoints to support further research.
 \clearpage
% Intentionally empty section placeholder.

% Add bibliography entries to custom.bib. Uncomment the next line once you cite a paper.
% Reference entries keep the default last-line behaviour
\setlength{\parfillskip}{0pt plus 1fil}
\bibliography{custom}

\clearpage
\appendix
% TODO: Add appendix sections if needed.

\section{Policy Optimization Details}
\label{app:policy-details}

For response $s$ recorded during RL, let $c_s$ be its actual model input and $y_{s,j}$ its $j$-th generated token. Its current trainer probability is
$p_{\theta,sj}=\pi_\theta(y_{s,j}\mid c_s,y_{s,<j})$.
Let $q_{sj}$ be its probability recomputed by the trainer under the round's starting weights, and let $b_{sj}$ be its probability recorded by the rollout server. All three probabilities condition on the same recorded context and response prefix. We use
\begin{equation}
\begin{aligned}
\rho_{sj}
&=\frac{p_{\theta,sj}}{q_{sj}},\\
w_{sj}
&=\operatorname{sg}\!\left[
\min\!\left(
\frac{q_{sj}}
{b_{sj}},C
\right)\right],
\end{aligned}
\label{ar:ratios}
\end{equation}
where $\operatorname{sg}$ denotes stop-gradient. The PPO ratio compares the current planner with the trainer's old-policy evaluation. We use truncated importance sampling~\citep{rolloutmismatch} to mitigate the probability mismatch between the rollout and training backends.

For ratio $\rho$ and advantage $A$, define
\begin{equation}
\begin{aligned}
\ell^{\mathrm{PPO}}&=\max\!\left\{-\rho A,\right.\\
&\qquad\left.-\operatorname{clip}(\rho,1-\epsilon_-,1+\epsilon_+)A\right\},\\
\ell_{\mathrm{DC}}(\rho,A)&=
\begin{cases}
\min\{\ell^{\mathrm{PPO}},-d A\},&A<0,\\
\ell^{\mathrm{PPO}},&A\geq0.
\end{cases}
\end{aligned}
\label{ar:dualclip}
\end{equation}
The main configuration uses $\epsilon_-=0.2$, $\epsilon_+=0.28$, $d=3$, and $C=2$. Equation~\eqref{ar:rl} minimizes this loss over response tokens; prompt and tool-result tokens do not receive policy-loss terms. The old-policy probabilities and importance weights remain fixed during the round's update. In the RL loss, response losses are token-averaged, weighted by inverse selected-turn count, and normalized within the optimizer batch. Because a batch need not contain complete episodes, this is not an exact episode-average objective. No KL or entropy term is added.

\section{Harness Adapters}
\label{app:adapters}

Table~\ref{tab:adapter-fields} lists the five fields an adapter reports for every episode and where each value comes from in our two harnesses. Integrating another harness requires producing these fields, routing the model calls of the planner through the recording proxy, and matching each recorded call to its episode. Sample construction and optimization are then shared.

\begin{table*}[t]
\centering
\small
\begin{tabular}{p{0.08\textwidth}p{0.26\textwidth}p{0.27\textwidth}p{0.27\textwidth}}
\hline
\textbf{Field} & \textbf{Meaning} & \textbf{RPent} & \textbf{CaP-X} \\
\hline
\texttt{task} & Task identifier & Suite and task index & Suite and task index \\
\texttt{init} & Initial-state identifier; with the task it defines the group & Initial-state seed & Trial index, which fixes the initial state \\
\texttt{attempt} & Repeat index within the group & Attempt index & Attempt index \\
\texttt{reward} & Terminal reward & Simulator success check, not the planner's own finish call & Task-completion flag written by the harness \\
\texttt{valid} & False for infrastructure failures, which are excluded from optimization & False when the failure is attributed to infrastructure & False for infrastructure failures. Errors in the generated code and unsuccessful execution are valid failures with zero reward \\
\hline
\end{tabular}
\caption{\textbf{Fields reported by a harness adapter and their sources in RPent and CaP-X.}}
\label{tab:adapter-fields}
\end{table*}

\section{Training Details}
\label{app:training}

\subsection{Supervised Fine-Tuning with RPent}

We fine-tune the 9B planner with AdamW, a learning rate of $10^{-5}$, and a batch of 64 target turns. We use $\beta_1 = 0.9$, $\beta_2 = 0.999$, $\epsilon = 10^{-8}$, and no weight decay. The learning rate rises over 10 warmup steps and then follows a cosine schedule that ends at 10\% of its initial value. Gradients are clipped to a global norm of 1.0. The language model and the visual merger are updated, and the vision encoder stays frozen.

The training set contains 40{,}047 target turns, so one epoch corresponds to 626 optimizer steps. The main results use the checkpoint at step 563, which is selected among ten checkpoints on the validation states 45--49 and is also the starting point for RL.

\subsection{Reinforcement Learning with RPent}

We run nine RL rounds and report the checkpoint after round 6. In each round the planner weights are fixed while rollouts are collected. A round covers 64 groups, where a group is one task with one initial state, and each group is attempted six times, which gives 384 episodes. Rollouts are sampled at temperature 1.0 without truncation of the token distribution. After outcomes and recorded model calls are validated, training samples are built and the planner is updated with several optimizer steps before the next round. We use a learning rate of $10^{-6}$ and a batch of 128 selected responses. At most 16 responses are selected from each episode. This limit applies to training samples and does not restrict the number of turns during execution. Groups in which all six attempts receive the same reward have zero advantage and contribute no training sample. Over the first six rounds, 2{,}304 episodes are collected, 26{,}052 responses are selected for optimization, and the planner receives 206 optimizer steps. The reported checkpoint is selected on the validation states 45--49, which are disjoint from the training and evaluation states.

\subsection{Training with CaP-X}

For CaP-X the teacher is the 27B planner in the same harness configuration as the student. We collect teacher episodes on the 26 tasks where the teacher solves at least one of the five validation episodes (initial states 45--49), and obtain 979 successful episodes with 2{,}370 target turns. Supervised fine-tuning uses the same learning rate, batch size, and warmup as with RPent and runs for one epoch, which is 37 optimizer steps. RL starts from this checkpoint and uses the same configuration as with RPent. Each round again covers 64 groups with six attempts. The groups are drawn from a pool of 35 tasks, which are the tasks where the SFT planner solves between one and four of the five validation episodes. Tasks that it always or never solves are left out because their groups would receive identical rewards. Rollouts on these tasks use initial states 11--44. The 35 tasks include all 26 tasks with teacher data and nine further tasks. The remaining 25 evaluation tasks are not used in training. Of the 35 tasks, 20 belong to the Object suites, nine to Goal, and six to Spatial. The reported planner is the checkpoint after 15 rounds.

\subsection{Episode Limits and Sampling Settings}

With RPent an episode ends after at most 100 planner turns or 7{,}200 seconds. With CaP-X a single response is limited to 32{,}768 tokens. The same limits are used for teacher collection, RL rollouts, and evaluation. Evaluation with RPent samples at temperature 1.0 with top-p 0.95, top-k 20, and a presence penalty of 1.5.

\end{document}